\documentclass[11pt]{article}
\usepackage{coling2020}
\usepackage{times}
\usepackage{url}
\usepackage{latexsym}
\usepackage{graphicx}
\colingfinalcopy 

\newcommand{\comment}[1]{}

\title{Discussion Tracker: 
Supporting Teacher Learning
about Students' Collaborative Argumentation in High School Classrooms}

\author{Luca Lugini \and Christopher Olshefski \and Ravneet Singh \and\\ {\bf Diane
  Litman} \and {\bf Amanda Godley} \\
         \\ University of Pittsburgh \\ Pittsburgh, PA, USA}

\date{}

\begin{document}
\maketitle
\begin{abstract}
Teaching collaborative argumentation is an advanced skill that many K-12 teachers struggle to develop. To address this, we have developed Discussion Tracker, a classroom discussion analytics system based on novel 
algorithms for 
classifying 
argument moves, specificity, and collaboration. Results from a classroom deployment indicate that teachers found the analytics useful, and that the underlying  classifiers perform with moderate to substantial agreement with humans.
\end{abstract}

%
%
\blfootnote{
    %
    %
    %
    %
    \hspace{-0.65cm}  
    This work is licensed under a Creative Commons 
    Attribution 4.0 International License.
    License details:
    \url{http://creativecommons.org/licenses/by/4.0/}.
}


\section{Introduction}
\label{intro}

Collaborative  argumentation 
in student dialogue 
is essential to individual learning as well as group problem-solving~\cite{Reznitskaya:13}.
Strong collaborative argumentation is characterized by specific claims, supporting evidence, and reasoning about that evidence as well as by building upon, questioning, and debating ideas posed by others. 
However, teaching collaborative argumentation is an advanced skill that many high school teachers struggle to develop \cite{lampert2010using},  
partially due to the practical challenge of keeping track of important features of students' talk while managing class and reflecting on students' talk when no record of it exists.  

To address this challenge, 
we have developed Discussion Tracker (DT), a system  that leverages natural language processing (NLP) to provide teachers with automatically
generated data about three important dimensions of students' collaborative argumentation: argument moves, specificity and collaboration. Discussion Tracker includes visualizations, interactive coded transcripts, collaboration maps, 
analytics across discussions, and 
instructional planning.  In contrast to teacher dashboards which largely focus on discussion analytics such as  amount 
of student/teacher talk, teacher wait time, and teacher question type~\cite{Chen:14,gerritsen2018towards,pehmer2015teacher,Blanchard:16}, DT focuses on students' collaborative argumentation.  In contrast to related NLP algorithms which largely focus on coding 
student essays~\cite{ghosh,klebanov,nguyen}, asynchronous online discussions~\cite{Swanson:15}, and news articles~\cite{Li:15}, DT's NLP algorithms address the challenges of coding transcripts of synchronous, face-to-face classroom discussions.


\section{Description of Discussion Tracker (DT)}
\label{sec:dt}

To use DT, a teacher first 
uploads a classroom discussion transcript. Next, NLP classifiers 
code the transcript using a previously developed scheme for representing three important dimensions of collaborative argumentation~\cite{lugini2018annotating,lrec2020}: argument moves (claim, evidence, explanation), specificity (low, medium, high), and collaboration (new, agree, extension, challenge/probe).  Student turns are the unit of analysis for collaboration. Argumentative Discourse Units (ADUs) ---  either entire turns, or segments within turns --- are the  argumentation and specificity units of analysis.

Each NLP classifier in DT was  developed by training on a previously collected and freely available corpus\footnote{\url{http://discussiontracker.cs.pitt.edu/} - we refer to this as corpus C1.} of collaborative argumentation~\cite{lrec2020}
using transformer-based neural networks.
A pretrained BERT model \cite{devlin2018bert,HuggingFace} is used to generate word embeddings for each word in an ADU (or turn, for collaboration). An average pooling layer is then used to compute the final embedding for the target ADU.
For predicting specificity, a softmax classifier is applied to the target ADU embedding.
For predicting argument moves, the target ADU as well as a window of surrounding ADUs are embedded, then
concatenated to form the final feature vector.
A softmax layer is applied on top of the feature vector to complete the argument move classifier.
This improves our prior argumentation models \cite{lugini2018argument} by using a pre-trained neural network and adding context information \cite{coling2020}.
The collaboration classifier is slightly more complex since collaboration labels depend on the relationship between a target turn and a particular reference turn.
For the purpose of this work we assume that the target turn is already provided in the input transcript.
A pretrained BERT model and average pooling layer are used to generate embeddings for the target  and  reference turns. An element-wise multiplication between the two embeddings is performed, yielding the feature vector used by a softmax classifier.

All models use the \textit{bert-base-uncased} BERT variant from the HuggingFace \cite{HuggingFace} library, which results in the smallest available dimensionality to keep computational complexity to a minimum.
The three models were built using the Keras library \cite{chollet2015keras}. The \textit{Adam} optimizer was used, as well as early stopping to automatically determine the number of epochs for training by monitoring validation loss (the validation set was chosen randomly and consisted of 10\% of the initial training set for each fold).


\begin{figure*}[t]
\centering
  \includegraphics[scale=0.25]{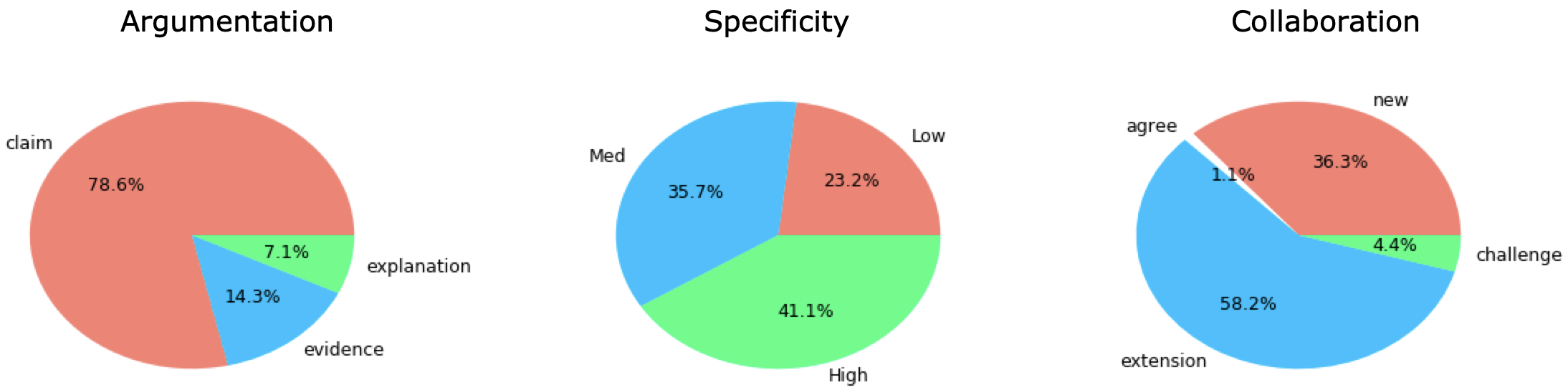}
\caption{Partial screenshot of ``Overview'' page in Discussion Tracker.}
\label{figure:overview}
\end{figure*}

\begin{figure}[t]
\centering
  \includegraphics[scale=0.25]{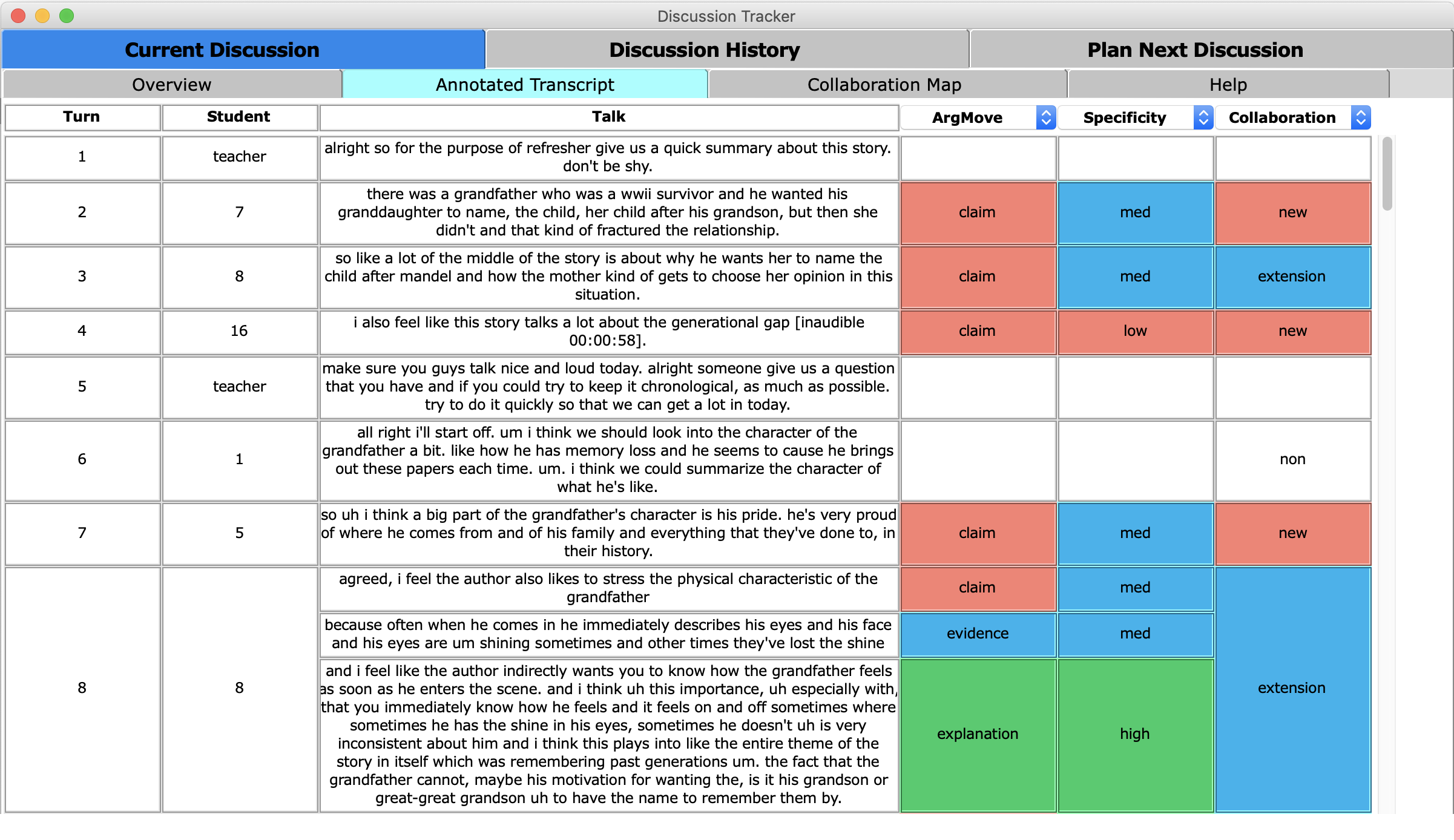}
\caption{Screenshot of ``Annotated Transcript'' page in Discussion Tracker.}
\label{figure:transcript}
\vspace{-2mm}
\end{figure}

\begin{figure}[t]
\centering
 \includegraphics[scale=0.25]{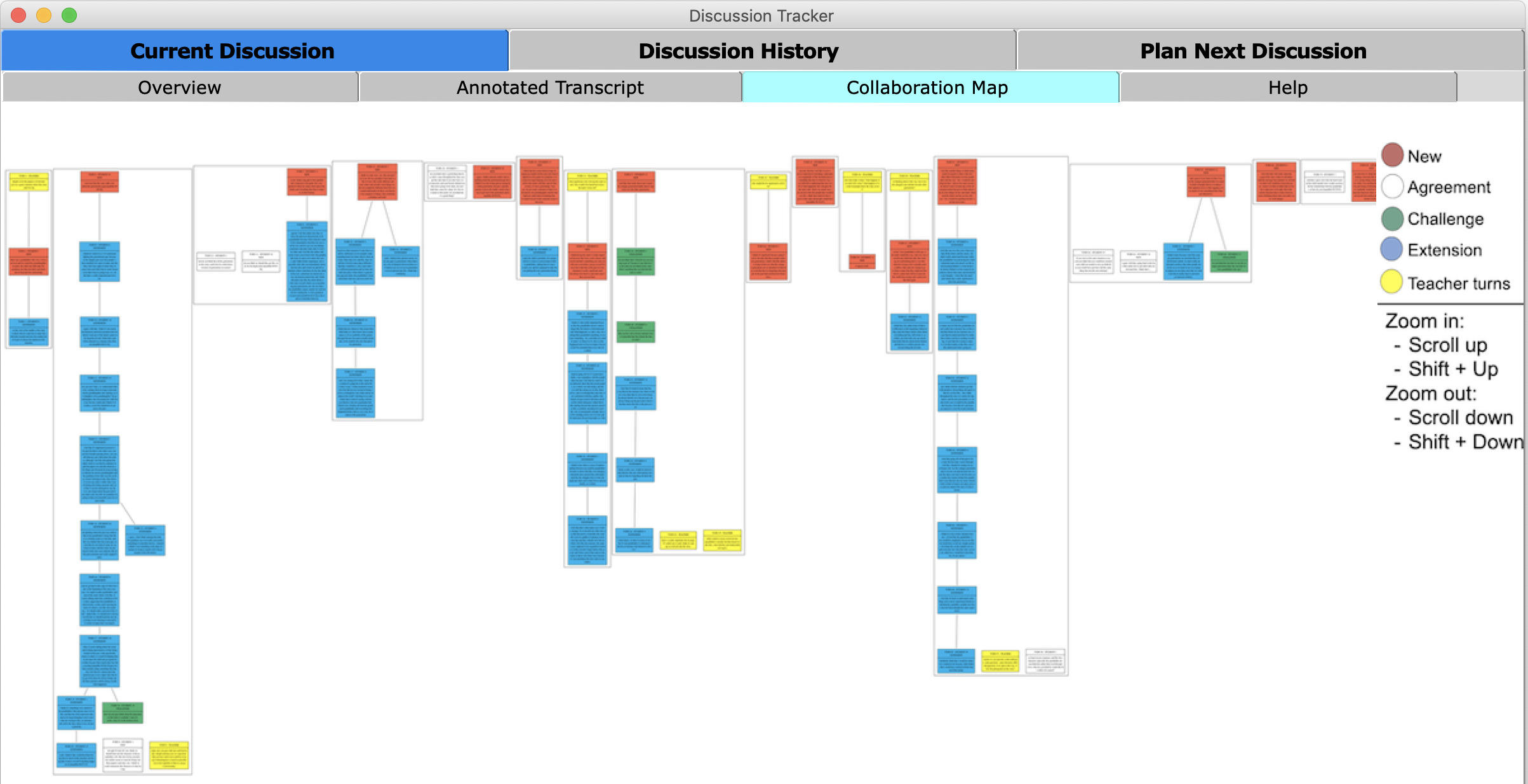}
\caption{Screenshot of ``Collaboration Map'' page in Discussion Tracker.}
\label{figure:map}
\vspace{-2mm}
\end{figure}

\begin{figure}[htb]
\centering
 \includegraphics[scale=0.25]{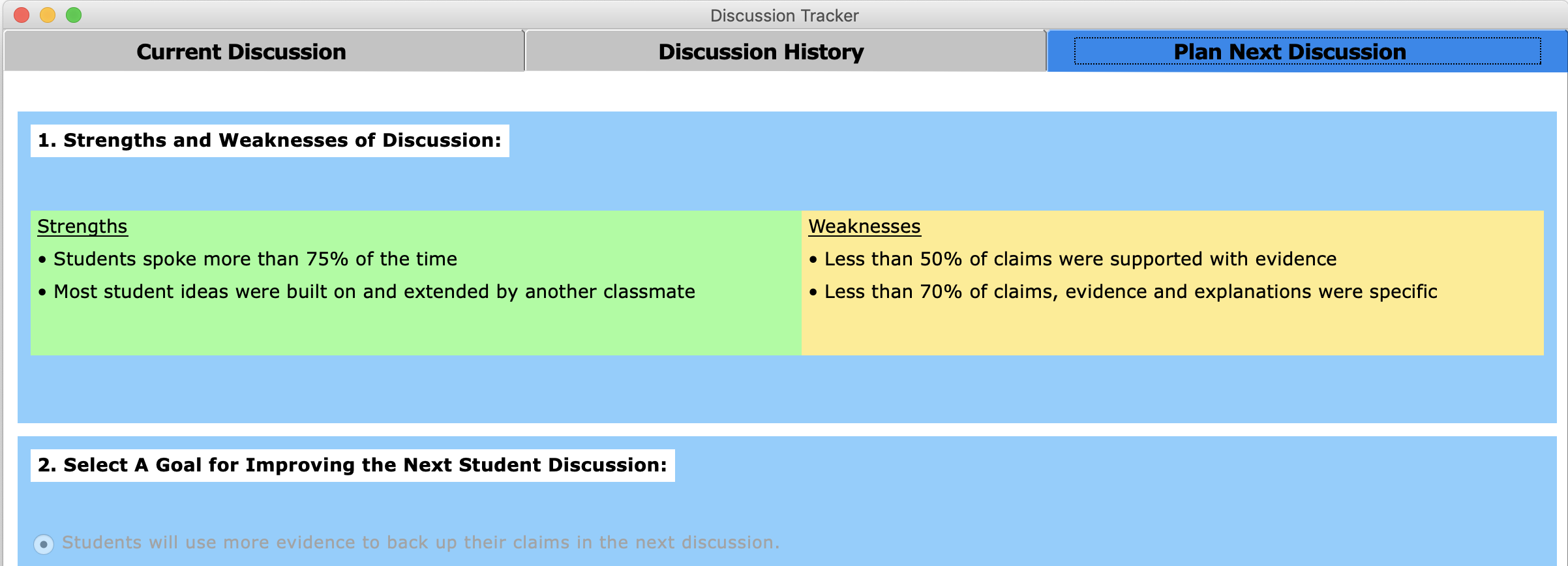}
\caption{Partial screenshot of ``Plan Next Discussion'' page in Discussion Tracker.}
\label{figure:goals}
\vspace{-2mm}
\end{figure}

After classification, all discussion analytics are automatically generated from the NLP codes. The DT overview screen (Figure \ref{figure:overview})  includes 
pie charts indicating the distribution of the codes for students' argument moves, specificity, and collaboration.  Other screens include interactive coded transcripts (Figure \ref{figure:transcript}), collaboration maps (Figure \ref{figure:map}), identification of strengths and weaknesses to support teacher goal-setting (Figure \ref{figure:goals}), and a history page (not shown) that compares the  code distributions
across discussions. 

We initially implemented a desktop version of DT using Python and Tkinter. The screenshots in the figures and the usability evaluation below are based on this version. 
To make DT more portable across hardware and to allow teachers to easily use DT on multiple machines (e.g., school, home), 
we now have  a web version of DT\footnote{Web app demo link and details and source at \url{discussiontracker.cs.pitt.edu}}. 
This version is implemented in Python and uses the REMI package\footnote{\url{https://github.com/dddomodossola/remi}} to convert Python into HTML and launch a webserver to accept requests for the site and handle user input.
With this setup it is easy to integrate the classifiers, implemented as a REST API on the same server hosting .

\section{Evaluation}

From January to March 2020, we collected data (corpus C2) to evaluate both teacher perceptions of DT as well as NLP classifier performance. In particular, the desktop version of DT was used by 18 high school English Language Arts  teachers from 4 schools, where:   1) each teacher led a  discussion about a literary text that was audio-recorded and observed by a researcher,  2) each teacher completed an online survey within a day, 
3) experienced annotators\footnote{ Kappa for argumentation (0.971) and collaboration (0.578) and Quadratic Weighted Kappa for specificity (0.813).}  hand-coded transcripts of the discussion for the three dimensions of collaborative argumentation discussed above and uploaded them into the DT system, 
4) within two weeks, researchers conducted a 45-minute cognitive interview \cite{voet2017history} with each teacher while they were using DT to look at their students' discussion\footnote{Teachers navigated DT with minimal training (a 15-minute, face-to-face demo).}, and 5) the same day, teachers completed a second survey that mirrored the first with additional items for ratings of DT.

{\bf DT Usability.}
We measured teachers' perceptions of the overall usefulness of DT and of specific features/visualizations through Likert-scale items on the survey from step 5 above.  Survey items were based on Holden and Rada’s \shortcite{holden2011understanding} teacher survey of perceived usability of technology. 
To remove noise that might distract from this usability evaluation, we evaluated DT under the best possible NLP conditions by using the manual codings of collaborative argumentation from step 3 above to generate all analytics. The NLP codings are separately evaluated in the classifier discussion below. 
Table~\ref{tab:survey}
indicates that teachers perceived DT to be very helpful for their learning about facilitating collaborative argumentation.
For nine of the 13  items, all teachers selected either ``Agree'' or ``Strongly agree'' (a mean score of 4.5), and no item received a ``Strongly disagree.'' 
Although the item ``I find the system easy to use'' received the lowest score (4.11), all teachers either agreed or agreed strongly with the item. Other items that scored higher, however, varied more in  responses. For example, although the majority of teachers agreed with ``The collaboration diagram is helpful,'' three neither agreed nor disagreed. 


\begin{table}
\centering
\begin{tabular}{|l|l|l|l|}
\hline
{\bf Question} & {\bf Mean} & {\bf Question} & {\bf Mean} \\ \hline
\begin{tabular}[c]{@{}l@{}}The overview of the discussion is  helpful.\end{tabular} & 4.67 & I find the system easy to use. & 4.11 \\ \hline
\begin{tabular}[c]{@{}l@{}}The pie charts of different features \\ of the student discussion are helpful.\end{tabular} & 4.78 & \begin{tabular}[c]{@{}l@{}}The system helps me to recognize \\ my students’ strengths during \\ discussion.\end{tabular} & 4.72 \\ \hline
\begin{tabular}[c]{@{}l@{}}The annotated transcript of student \\ discussion is helpful.\end{tabular} & 4.89 & \begin{tabular}[c]{@{}l@{}}The system helps me to recognize \\ my students’ weakness during \\ discussion.\end{tabular} & 4.72 \\ \hline
\begin{tabular}[c]{@{}l@{}}The collaboration diagram is  helpful.\end{tabular} & 4.22 & \begin{tabular}[c]{@{}l@{}}The system gives me more insight\\  into student learning than I usually get \\ from thinking about the discussion.\end{tabular} & 4.67 \\ \hline
\begin{tabular}[c]{@{}l@{}}The system-generated strengths \\ and weaknesses are helpful.\end{tabular} & 4.44 & \begin{tabular}[c]{@{}l@{}} The system encourages me to \\ make more changes to my facilitation \\ of discussion than I usually do.\end{tabular} & 4.28 \\ \hline
The goal-setting is helpful. & 4.56 & \begin{tabular}[c]{@{}l@{}}Overall, Discussion Tracker is \\ helpful for my teaching of literature \\ discussions.\end{tabular} & 4.72 \\ \hline
\begin{tabular}[c]{@{}l@{}} The instructional resources are  helpful.\end{tabular} & 4.17 &  &  \\ \hline
\end{tabular}
\caption{Teacher survey items and Likert score means.}
\label{tab:survey}
\end{table}

{\bf NLP Classifier Performance.} As the gold standard for evaluating DT classifier performance, we used the manual annotations from step 3 of the data collection discussed above. 
Table~\ref{tab:stats} shows the distribution of the gold-standard codes, while Table~\ref{tab:results} shows classifier performance when compared to these gold-standards.\footnote{The input for the gold-standard and automated coding was identical (loosely, a spreadsheet version of the first three columns in Figure~\ref{figure:transcript}).
A professional service (rather than ASR) performed the audio transcription and segmentation into  turns  (for collaboration coding).  A researcher further segmented student turns into ADUs (for argumentation and specificity coding).}
\begin{table}[!t]
\centering
\begin{tabular}{|l|l|}
\hline
{\bf Annotation} &{\bf Distribution} \\ \hline
Argumentation & claim (72\%), evidence (18\%), explanation (10\%) \\ \hline
Specificity & low (29\%), medium (36\%), high (36\%) \\ \hline
Collaboration & new (22\%), agree (3\%), extensions (54\%), challenge/probe (21\%) \\ \hline
\end{tabular}
\caption{Descriptive statistics of gold-standard annotations in test corpus.}
\label{tab:stats}
\end{table} 
\begin{table}[t!]
\centering
\begin{tabular}{|c|c|c|c|c|c|} \hline
\textbf{Code}  &\textbf{N} & \textbf{Kappa} & \textbf{Macro F} &\textbf{Micro F} \\ \hline
Argument Move & 1942 ADUs & 0.574 & 0.730 & 0.789  \\ \hline
Specificity & 1942 ADUs & 0.727 & 0.688 & 0.679  \\ \hline
Collaboration & 1467 Turns & 0.566 & 0.439 & 0.775 \\ \hline
\end{tabular}
\caption{Transfomer-based neural classification results. }
\label{tab:results}
\vspace{-2mm}
\end{table}
The results in Table~\ref{tab:results} were obtained by training each classifier separately on corpus C1 (footnote 1) and testing on corpus C2 (the 18 discussions collected in this study).
Hyperparameter optimization was performed using cross-validation on C1 in order to find out how much contextual information before/after the target ADU to consider (i.e. context window size).
This yielded an argument classifier 
that added a window of 2 ADUs preceding and 2 ADUs following the target ADU for embedding.
Though all classifiers show respectable results, predictions for argument move and specificity are more consistent  for individual class labels, as evidenced by the small difference between macro and micro F-score.
The lower macro F-score for collaboration is due to  poor prediction performance for the agree and challenge/probe codes.

\section{Summary and Future Directions}
In this work we described the development of a classroom analytics system and reported usability results from real world classroom deployment. We conducted a survey that showed teachers found the system easy to use and the analytics (based on human-annotated labels) helpful in analyzing collaborative argumentation.
Evaluation of the automated NLP classifiers showed that they are in moderate to substantial agreement with the labels provided by human annotators.
The main goal of future work is to continue to enhance our neural classification methods, and to develop an end-to-end, completely automated system.
To this end, we will consider several aspects: perform new data collections and improve classifier performance; incorporate Automatic Speech Recognition to perform automated transcription; develop algorithms to automatically segment turns into ADUs.
In addition, we will further develop the interface by addressing teacher feedback and improving the system's ease of use.  Finally,  teachers will evaluate our newer versions of DT, including versions where the analytics are based on classifier outputs rather than human-annotated labels. 


\section*{Acknowledgements}

This work was supported by the National Science Foundation (1842334 and 1917673) and by the University of Pittsburgh's Learning Research and Development Center as well as Center for Research Computing
through the resources provided. We would like to thank the teachers who participated in this study.

\bibliographystyle{coling}
\bibliography{references,colingRefs}

\begin{thebibliography}{}

\bibitem[\protect\citename{Blanchard \bgroup et al.\egroup }2016]{Blanchard:16}
Nathaniel Blanchard, Patrick~J Donnelly, Andrew~M Olney, Borhan Samei, Brooke
  Ward, Xiaoyi Sun, Sean Kelly, Martin Nystrand, and Sidney~K. D'Mello.
\newblock 2016.
\newblock Identifying teacher questions using automatic speech recognition in
  classrooms.
\newblock In {\em 17th Annual Meeting of the Special Interest Group on
  Discourse and Dialogue}, page 191.

\bibitem[\protect\citename{Chen \bgroup et al.\egroup }2014]{Chen:14}
G~Chen, SN~Clarke, and LB~Resnick.
\newblock 2014.
\newblock An analytic tool for supporting teachers’ reflection on classroom
  talk.
\newblock In {\em Learning and becoming in practice: The International
  Conference of the Learning Sciences (ICLS) 2014}. International Society of
  the Learning Sciences.

\bibitem[\protect\citename{Chollet and others}2015]{chollet2015keras}
Fran\c{c}ois Chollet et~al.
\newblock 2015.
\newblock Keras.
\newblock \url{https://keras.io}.

\bibitem[\protect\citename{Devlin \bgroup et al.\egroup }2019]{devlin2018bert}
Jacob Devlin, Ming-Wei Chang, Kenton Lee, and Kristina Toutanova.
\newblock 2019.
\newblock {BERT}: Pre-training of deep bidirectional transformers for language
  understanding.
\newblock In {\em Proceedings of the 2019 Conference of the North {A}merican
  Chapter of the Association for Computational Linguistics: Human Language
  Technologies}, pages 4171--4186, Minneapolis, Minnesota, June.

\bibitem[\protect\citename{Gerritsen \bgroup et al.\egroup
  }2018]{gerritsen2018towards}
David Gerritsen, John Zimmerman, and Amy Ogan.
\newblock 2018.
\newblock Towards a framework for smart classrooms that teach instructors to
  teach.
\newblock In {\em International Conference of the Learning Sciences}, volume~3.

\bibitem[\protect\citename{Ghosh \bgroup et al.\egroup }2016]{ghosh}
Debanjan Ghosh, Aquila Khanam, Yubo Han, and Smaranda Muresan.
\newblock 2016.
\newblock Coarse-grained {Argumentation} {Features} for {Scoring} {Persuasive}
  {Essays}.
\newblock In {\em Proceedings of the 54th {Annual} {Meeting} of the
  {Association} for {Computational} {Linguistics} ({Volume} 2: {Short}
  {Papers})}, pages 549--554.

\bibitem[\protect\citename{Holden and Rada}2011]{holden2011understanding}
Heather Holden and Roy Rada.
\newblock 2011.
\newblock Understanding the influence of perceived usability and technology
  self-efficacy on teachers’ technology acceptance.
\newblock {\em Journal of Research on Technology in Education}, 43(4):343--367.

\bibitem[\protect\citename{Klebanov \bgroup et al.\egroup }2016]{klebanov}
Beata~Beigman Klebanov, Christian Stab, Jill Burstein, Yi~Song, Binod Gyawali,
  and Iryna Gurevych.
\newblock 2016.
\newblock Argumentation: {Content}, {Structure}, and {Relationship} with
  {Essay} {Quality}.
\newblock In {\em Proceedings of the {Third} {Workshop} on {Argument} {Mining}
  ({ArgMining}2016)}, pages 70--75, Berlin, Germany.

\bibitem[\protect\citename{Lampert \bgroup et al.\egroup
  }2010]{lampert2010using}
Magdalene Lampert, Heather Beasley, Hala Ghousseini, Elham Kazemi, and Megan
  Franke.
\newblock 2010.
\newblock Using designed instructional activities to enable novices to manage
  ambitious mathematics teaching.
\newblock In {\em Instructional explanations in the disciplines}, pages
  129--141. Springer.

\bibitem[\protect\citename{Li and Nenkova}2015]{Li:15}
Junyi~Jessy Li and Ani Nenkova.
\newblock 2015.
\newblock Fast and accurate prediction of sentence specificity.
\newblock In {\em Proceedings of the Twenty-Ninth Conference on Artificial
  Intelligence (AAAI)}, pages 2281--2287, January.

\bibitem[\protect\citename{Lugini and Litman}2018]{lugini2018argument}
Luca Lugini and Diane Litman.
\newblock 2018.
\newblock Argument component classification for classroom discussions.
\newblock In {\em Proceedings of the 5th Workshop on Argument Mining}, pages
  57--67.

\bibitem[\protect\citename{Lugini and Litman}2020]{coling2020}
Luca Lugini and Diane Litman.
\newblock 2020.
\newblock Contextual argument component classification for class discussions.
\newblock In {\em Proceedings s of the 28th International Conference on
  Computational Linguistics}, Online, December.

\bibitem[\protect\citename{Lugini \bgroup et al.\egroup
  }2018]{lugini2018annotating}
Luca Lugini, Diane Litman, Amanda Godley, and Christopher Olshefski.
\newblock 2018.
\newblock Annotating student talk in text-based classroom discussions.
\newblock In {\em Proceedings of the Thirteenth Workshop on Innovative Use of
  NLP for Building Educational Applications}, pages 110--116.

\bibitem[\protect\citename{Nguyen and Litman}2016]{nguyen}
Huy Nguyen and Diane Litman.
\newblock 2016.
\newblock Context-aware {Argumentative} {Relation} {Mining}.
\newblock In {\em Proceedings of the 54th {Annual} {Meeting} of the
  {Association} for {Computational} {Linguistics} ({Volume} 1: {Long}
  {Papers})}, pages 1127--1137, Berlin, Germany.

\bibitem[\protect\citename{Olshefski \bgroup et al.\egroup }2020]{lrec2020}
Christopher Olshefski, Luca Lugini, Ravneet Singh, Diane Litman, and Amanda
  Godley.
\newblock 2020.
\newblock The {D}iscussion {T}racker corpus of collaborative argumentation.
\newblock In {\em Proceedings of the 12th International Conference on Language
  Resources and Evaluation}, pages 1033--1043, Marseille, France, May.

\bibitem[\protect\citename{Pehmer \bgroup et al.\egroup
  }2015]{pehmer2015teacher}
Ann-Kathrin Pehmer, Alexander Gr{\"o}schner, and Tina Seidel.
\newblock 2015.
\newblock How teacher professional development regarding classroom dialogue
  affects students' higher-order learning.
\newblock {\em Teaching and Teacher Education}, 47:108--119.

\bibitem[\protect\citename{Reznitskaya and Gregory}2013]{Reznitskaya:13}
Alina Reznitskaya and Maughn Gregory.
\newblock 2013.
\newblock Student thought and classroom language: Examining the mechanisms of
  change in dialogic teaching.
\newblock {\em Educational Psychologist}, 48(2):114--133.

\bibitem[\protect\citename{Swanson \bgroup et al.\egroup }2015]{Swanson:15}
Reid Swanson, Brian Ecker, and Marilyn Walker.
\newblock 2015.
\newblock Argument mining: Extracting arguments from online dialogue.
\newblock In {\em Proceedings of the 16th Annual Meeting of the Special
  Interest Group on Discourse and Dialogue}, pages 217--226.

\bibitem[\protect\citename{Voet and Wever}2017]{voet2017history}
Michiel Voet and Bram~De Wever.
\newblock 2017.
\newblock History teachers’ knowledge of inquiry methods: An analysis of
  cognitive processes used during a historical inquiry.
\newblock {\em Journal of Teacher Education}, 68(3):312--329.

\bibitem[\protect\citename{Wolf \bgroup et al.\egroup }2019]{HuggingFace}
Thomas Wolf, Lysandre Debut, Victor Sanh, Julien Chaumond, Clement Delangue,
  Anthony Moi, Pierric Cistac, Tim Rault, R'emi Louf, Morgan Funtowicz, and
  Jamie Brew.
\newblock 2019.
\newblock Huggingface's transformers: State-of-the-art natural language
  processing.
\newblock {\em ArXiv}, abs/1910.03771.

\end{thebibliography}

\end{document}